\documentclass{article}

\usepackage{PRIMEarxiv}

\usepackage[utf8]{inputenc} % allow utf-8 input
\usepackage[T1]{fontenc}    % use 8-bit T1 fonts
\usepackage{hyperref}       % hyperlinks
\usepackage{url}            % simple URL typesetting
\usepackage{booktabs}       % professional-quality tables
\usepackage{amsfonts}       % blackboard math symbols
\usepackage{nicefrac}       % compact symbols for 1/2, etc.
\usepackage{microtype}      % microtypography
\usepackage{lipsum}
\usepackage{fancyhdr}       % header
\usepackage{graphicx}       % graphics
\graphicspath{{media/}}     % organize your images and other figures under media/ folder
\usepackage{caption}

\usepackage{authblk} % 著者と所属を管理するパッケージ
\usepackage[whole]{bxcjkjatype}
\usepackage{float} % 浮動体の挙動を制御するパッケージ

\title{Heron-Bench: A Benchmark for Evaluating Vision Language Models in Japanese
}

\author[1,2,*,†]{Yuichi Inoue}
\author[1,3,†]{Kento Sasaki}
\author[1,4]{Yuma Ochi}
\author[1,5]{Kazuki Fujii}
\author[1]{Kotaro Tanahashi}
\author[1,2,*]{Yu Yamaguchi}

\affil[1]{Turing Inc.}
\affil[2]{Keio Research Institute at SFC}
\affil[3]{University of Tsukuba}
\affil[4]{National Institute of Technology, Kisarazu College}
\affil[5]{Tokyo Institute of Technology}
\affil[*]{\{y.inoue, yu.yamaguchi\}@turing-motors.com}
\affil[ ]{† Contributed equally to this work.}

\begin{document}

\maketitle

\begin{abstract}

Vision Language Models (VLMs) have undergone a rapid evolution, giving rise to significant advancements in the realm of multimodal understanding tasks. However, the majority of these models are trained and evaluated on English-centric datasets, leaving a gap in the development and evaluation of VLMs for other languages, such as Japanese. This gap can be attributed to the lack of methodologies for constructing VLMs and the absence of benchmarks to accurately measure their performance. To address this issue, we introduce a novel benchmark, Japanese Heron-Bench, for evaluating Japanese capabilities of VLMs. The Japanese Heron-Bench consists of a variety of image-question answer pairs tailored to the Japanese context. Additionally, we present a baseline Japanese VLM that has been trained with Japanese visual instruction tuning datasets. Our Heron-Bench reveals the strengths and limitations of the proposed VLM across various ability dimensions. Furthermore, we clarify the capability gap between strong closed models like GPT-4V~\cite{OpenAI_GPT4_2023, 2023GPT4VisionSC} and the baseline model, providing valuable insights for future research in this domain. We release the benchmark dataset and training code to facilitate further developments in Japanese VLM research.
\end{abstract}

% keywords can be removed
% \keywords{First keyword \and Second keyword \and More}

% ==============================================
% Figure
% ==============================================
% \begin{figure}[t!]
\begin{figure}[b!]
    \centering
    \includegraphics[width=400pt]{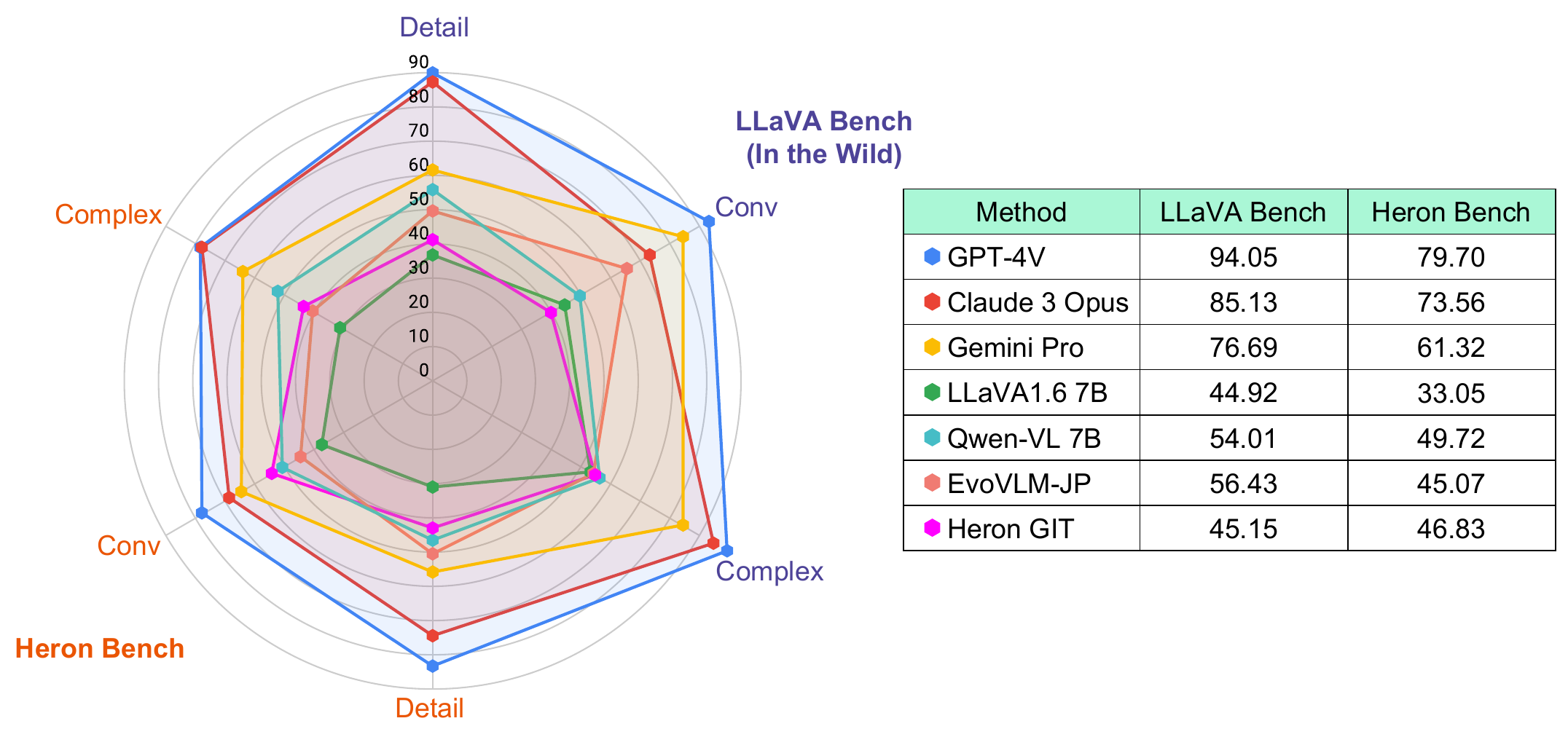}
    \caption{\textbf{Comparison of evaluation results using the Japanese-translated LLaVA Bench (In the Wild) and the Japanese Heron-Bench.}}
    \label{fig:lader_chart}
\end{figure}

\newpage

\section{Introduction}

Rapid advancements of Large Language Models (LLMs) mark a cornerstone in the development of artificial intelligence. Recently, various methods for developing LLMs have been proposed and well-trained models have become increasingly public. The development of LLMs is not limited to the English language; efforts have been made to build LLMs in other languages, including Japanese~\cite{swallow7b, OpenCalm7b, JapaneseStableLMBaseAlpha7B, StabilityGamma7b, RinnaNekomata7b, elyzallama2023}.

On the basis of the progress in LLMs, approaches have been proposed for the construction of Vision Language Models (VLMs), which extend LLMs with image encoders~\cite{NEURIPS2023_6dcf277e, liu2023improved, liu2024llavanext, bai2023qwenvl, blip2, dai2023instructblip, zhu2023minigpt4, fuyu_8b, lu2023empirical, li2024minigemini}. In addition to the developments of VLM's training, various evaluation metrics have been proposed to assess their performance, including image captioning evaluation metrics~\cite{papineni-etal-2002-bleu, conf_cvpr_VedantamZP15, banerjee-lavie-2005-meteor, lin-2004-rouge}, scoring similarity between images and text~\cite{hessel-etal-2021-clipscore}, and accuracy of visual question answering (VQA)~\cite{Goyal_2017_CVPR, Hudson_2019_CVPR, VizWiz}. Furthermore, recent studies~\cite{NEURIPS2023_6dcf277e, fu2023mme, liu2023mmbench, bai2023touchstone} have proposed more comprehensive methods specifically designed to evaluate state-of-the-art large VLMs, taking into account their ability to handle a wide range of tasks and their robustness to visual scene understanding. However, it is important to note that most of the current high-performing VLMs are trained predominantly on English-centric datasets, and evaluated by English datasets. With the rapid development and increasing popularity of VLMs, the demand for non-English models is growing, and there is an urgent need to accurately understand the capabilities of VLMs when applied to images deeply rooted in the cultural and linguistic contexts of each region. In the case of Japanese, training methods for large VLMs are not well-described, and even when models are released, the evaluation of these models remains insufficient.

In this work, we introduce a new evaluation benchmark, named Japanese Heron-Bench, for assessing the performance of VLMs in the Japanese language. This benchmark dataset consists of newly collected images and 102 questions unique to the Japanese context. Using this dataset, we can effectively analyze the abilities of VLMs to understand visual scenes and answer questions in the Japanese context. Furthermore, we introduce a method for constructing a Japanese VLM trained on Japanese image-text pairs using Japanese LLMs. The Japanese VLM developed in this study serves as a baseline for the proposed evaluation dataset. We make the training code, the trained model, and the evaluation dataset publicly available~\footnote{\url{https://github.com/turingmotors/heron}},~\footnote{\url{https://huggingface.co/turing-motors}}.

% ==============================================
% Related work
% ==============================================
\section{Related Work}

\subsection{VLM Evaluation Datasets}

Various methods have been proposed for evaluating VLMs. For image captioning evaluation, metrics such as BLEU~\cite{papineni-etal-2002-bleu}, ROUGE~\cite{lin-2004-rouge} and METEOR~\cite{banerjee-lavie-2005-meteor} are commonly used, which assess performance by measuring the similarity between generated and reference sentences based on n-grams. To measure the similarity between images and text, methods using CLIP and cosine similarity have been proposed~\cite{hessel-etal-2021-clipscore}. Furthermore, evaluation methods such as VQAv2~\cite{Goyal_2017_CVPR}, GQA~\cite{Hudson_2019_CVPR} and VizWiz\cite{VizWiz} have been developed to assess the accuracy of visual question answering. With the recent advancements in LLMs, more comprehensive evaluation methods for VLMs that demonstrate advanced language capabilities have also been proposed. LLaVA-Bench (COCO, In-the-Wild)~\cite{NEURIPS2023_6dcf277e} and TouchStone\cite{bai2023touchstone} leverage GPT-4~\cite{OpenAI_GPT4_2023} to directly score the sentences generated by the models. While these evaluation methods are well-established for English, few options are available for evaluating Japanese VLMs.

\subsection{Construction of VLMs}

There have been several approaches proposed for constructing VLMs, such as GIT~\cite{wang2022git} and BLIP~\cite{pmlr-v162-li22n}, which combine language models, image encoders, and adapters to connect them. With the recent advancements in LLMs, model architectures and training techniques have been proposed to leverage the text generation capabilities of LLMs to acquire high explanatory power for images. Flamingo~\cite{Flamingo} bridges pretrained vision and language models using a Perceiver Resampler to extract visual features and inject them into the language model using cross attention layers, enabling rapid adaptation to various tasks with few annotated examples. BLIP-2~\cite{blip2, dai2023instructblip} introduces a transformer-based module called Q-Former, which uses cross-attention to extract fixed-length query vectors from the image vectors obtained by the image encoder, creating image tokens that can be treated similarly to text tokens. LLaVA~\cite{NEURIPS2023_6dcf277e, liu2023improved, liu2024llavanext} obtains image tokens by passing the image vectors obtained from the image encoder through several several feed-forward networks, and then inputs these image tokens along with text tokens into the LLM. Moreover, they also introduced a fine-tuning method called visual instruction tuning, which takes advantage of the strong language capabilities of LLMs and aligns a VLM with human intent by using a smaller number of image-text pairs than required in pre-training.

% ==============================================
% Figure
% ==============================================
\begin{figure}[t!]
    \centering
    \includegraphics[width=450pt]{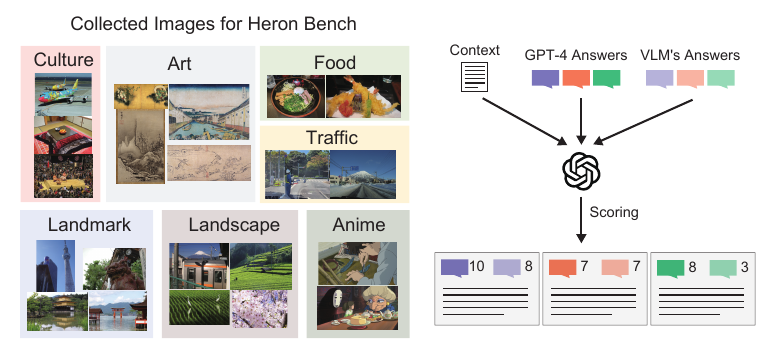}
    \caption{\textbf{Overview of collected images for evaluation per subcategory and scoring process.} The dataset consists of seven categories relevant to the Japanese context. The GPT-4 API is used to evaluate and score the answers provided by both the VLMs and GPT-4. The context works as a reference for scoring the answers.}
    \label{fig:method_overview}
\end{figure}

% ==============================================
% Methodology
% ==============================================
\section{Japanese Heron-Bench}

This section first describes the dataset construction and evaluation methods created to assess the image-description and question-answering abilities of VLMs in the Japanese context. Then, it explains the baseline model construction.

\subsection{Evaluation Method for Japanese VLMs}

The creation of the Heron-Bench evaluation set follows the construction method of LLaVA-Bench (In-the-Wild). An overview of the evaluation dataset and scoring method is shown in Figure~\ref{fig:method_overview}.

\paragraph{Dataset Construction}

For the evaluation, we collected 21 public domain or CC BY 2.0 licensed images related to Japan. We then set up three categories for each image: Conversation, Detail, and Complex, and prepared one or two questions for each category. The final evaluation dataset consists of 102 questions. Furthermore, each image is assigned one of seven subcategories: anime, art, culture, food, landscape, landmark, and transportation.

In order to create model answers, we manually describe the information about the image in detail as context. Then, we provide the context and questions to the GPT-4 API (gpt-4-0125-preview) to generate model answers for the questions, which are used for evaluation. (See also Appendix~\ref{appendix:a}.)

\paragraph{Scoring Method}

The score calculation is the same way as proposed in the LLaVA Bench~\cite{NEURIPS2023_6dcf277e}. First, the images and questions are input into the VLM to evaluate, and the answer texts are obtained. The obtained answers, the GPT-4's answers, and the contexts (ground truth) are then evaluated by using GPT-4 API. The GPT-4 API is instructed to assign scores out of 10 to both the GPT-4's answers and the VLM's answers based on the context and to provide explanations for the scores. The final VLM's score is determined by the ratio of the average score of the VLM's answers to the average score of the GPT-4 model answers.

\subsection{Baseline Model Construction}

To fully leverage this benchmark, we have trained a baseline model in a language-aware manner. This baseline model clarifies the current performance gap with high-performing closed models and serves as a reference point for future VLMs. For the model training, we adopted the visual instruction tuning method proposed for developing LLaVA-1.6~\cite{liu2023improved}. The dataset consists of approximately 558K samples used for pre-training the adapter, and about 665K image-text pair samples used for instruction tuning when the LLM and adapter parameters were unfrozen during training. Both datasets were translated into Japanese using the DeepL API. The 665K image-text pair dataset contains text only samples. In our experiments, we excluded the text-only samples and used only the about 620K image-text pair samples.

Regarding the model architecture, similar to GIT, we used a single linear layer as an adapter after the image encoder to convert image vectors into image tokens that can be treated similarly to text tokens. We employed OpenAI's CLIP Large Patch 14 (336)~\cite{OpenaiCLIP} as the image encoder and StabilityAI's japanese-stablelm-base-alpha-7b~\cite{JapaneseStableLMBaseAlpha7B} as the Japanese LLM. We used global batch sizes of 256 and 128 and learning rates of 1e-3 and 1e-5 for the first and second stages, respectively. For learning rate scheduling, we adopted a linear scheduler with a warmup period.

% ==============================================
% Experiments
% ==============================================
\section{Experiments}

\subsection{Evaluation Benchmarks}
For model evaluation, in addition to the proposed Heron-Bench, we also use LLaVA-Bench (COCO) and LLaVA-Bench (In-the-Wild), which were translated into Japanese using DeepL and manually modulated. We evaluate both open VLMs that are publicly available and closed VLMs that can be accessed via APIs. The following models were evaluated:

\begin{enumerate}
    \item[Closed: ] GPT-4V~\cite{OpenAI_GPT4_2023, 2023GPT4VisionSC}, Claude 3 Opus~\cite{claude3}, Gemini Pro Vision~\cite{reid2024gemini}
    \item[Open: ] Heron GIT (proposed in this paper), Heron BLIP v1, LLaVA-1.6\cite{liu2024llavanext}, LLaVA-1.5\cite{liu2023improved}, Qwen-VL\cite{bai2023qwenvl}, Japanese Stable VLM\cite{JapaneseStableVLM}, EvoVLM-JP\cite{akiba2024evolutionary}
\end{enumerate}

% ==============================================
% Table
% ==============================================
\begin{table}[t!]
    \centering
    \includegraphics[width=1.0\linewidth]{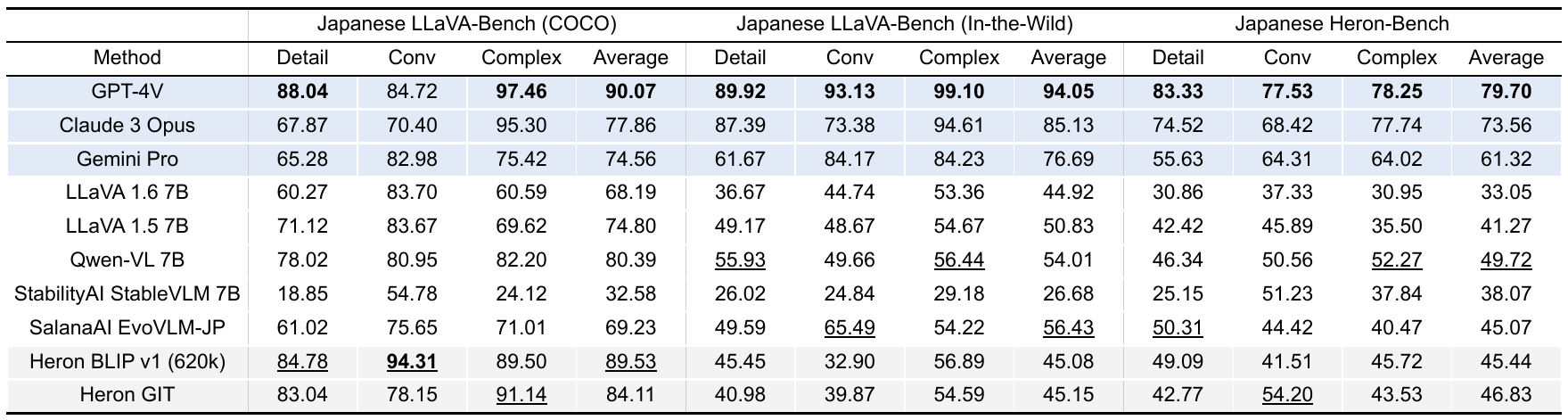}
    \caption{\textbf{Evaluation of VLMs using Japanese LLaVA-Bench (COCO), Japanese LLaVA-Bench (In-the-Wild), and Heron-Bench.} The blue background indicates closed VLMs accessible via APIs. The gray background represents VLMs we previously released (Heron BLIP v1) and the VLM we provided in this paper (Heron GIT).}
    \label{fig:Table}
\end{table}

\subsection{Quantitative Evaluation}

Table~\ref{fig:Table} shows the scores of all models evaluated in this paper. First we focus on results of the Japanese LLaVA-Bench (In-the-Wild) and the Japanese Heron-Bench. Closed models consistently achieve high scores, with GPT-4V performing exceptionally well across almost all evaluation metrics. Among the open models, Qwen-VL, which is trained on large-scale image-text pairs, consistently obtains high scores. Heron BLIP v1 and Heron GIT, which undergo instruction tuning using Japanese image-text pairs, achieve decent results on the Heron-Bench, but their scores are lower on the Japanese LLaVA-Bench (In-the-Wild). LLaVA-1.5 performs well on the LLaVA-Bench (In-the-Wild) in the English context, but its scores tend to decrease on the Heron-Bench, which is more heavily based on the Japanese context. LLaVA-1.6 exhibits lower scores on Japanese question-answering compared to English, suggesting that its Japanese language capability is not as advanced. The results of LLaVA-1.5 and LLaVA-1.6 are likely due to the limited amount of Japanese data in their training datasets. Interestingly, EvoVLM-JP, which is developed using evolutionary model merging, achieves a higher score on the Japanese LLaVA-Bench (In-the-Wild) than the other open models.

Open models show higher scores comparable to closed models in LLaVA-Bench (COCO). However, they tend to have significantly lower scores in LLaVA-Bench (In-the-Wild) and Heron-Bench. The qualitative evaluation in section~\ref{sec:qualitative} suggests that the actual capability gap between closed and open VLMs is closer to the score differences observed in LLaVA-Bench (In-the-Wild) and Heron-Bench. Therefore, LLaVA-Bench (COCO) may not be well-suited for measuring the Japanese language capabilities of VLMs. On the other hand, the proposed Heron-Bench, which maintains a similar difficulty level to LLaVA-Bench (In-the-Wild) while using images and questions related to Japan, is considered a useful benchmark for evaluating the Japanese language understanding capabilities of VLMs.

% =============================================================
% Subcategory Figures
% =============================================================
\begin{figure}[t!]
    % Left figure
    \begin{minipage}[t]{0.53\linewidth}
        \centering
        \includegraphics[width=\linewidth]{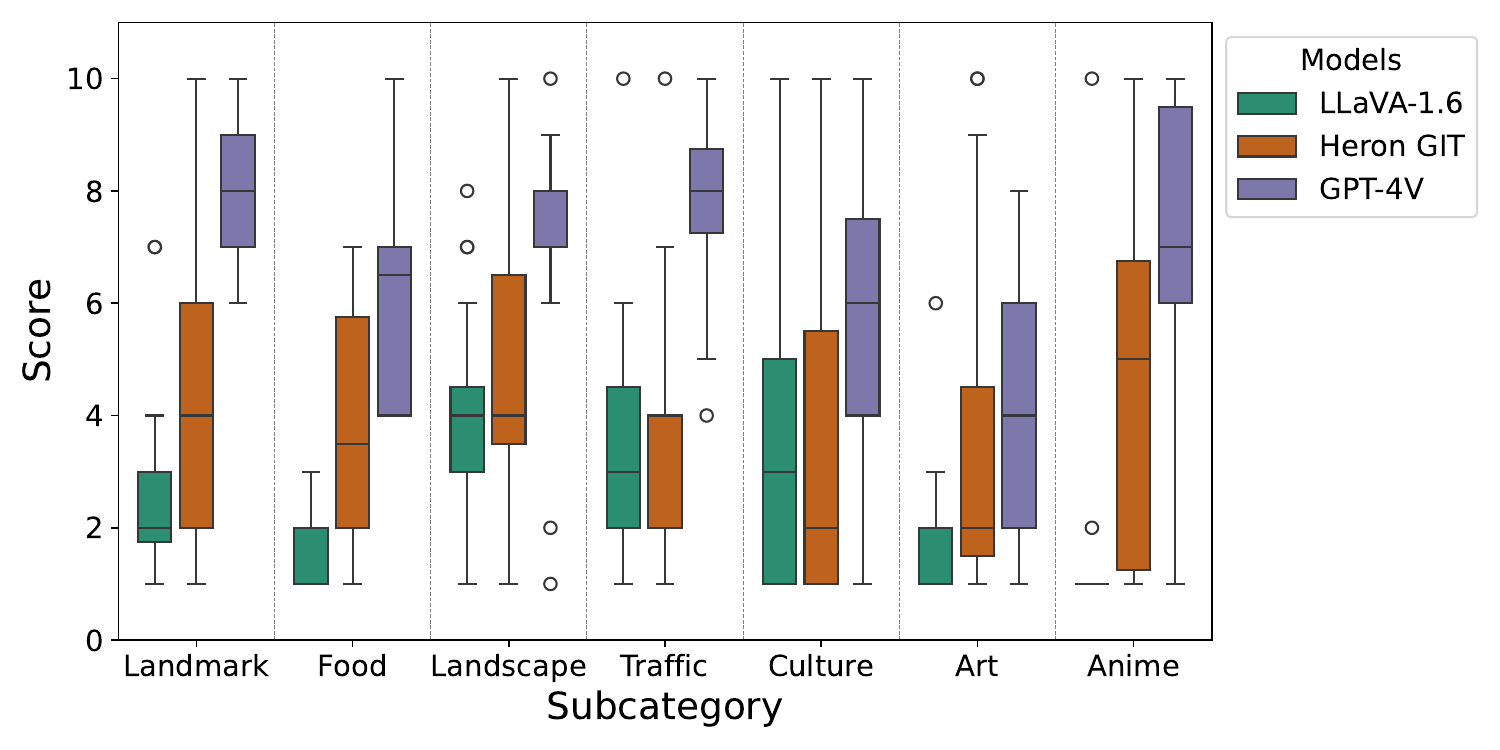} % Assuming the content of the table is an image
        \caption{\textbf{Comparison of scores of GPT-4V (closed model), Heron GIT (Japanese VLM), and LLaVA-1.6 (English VLM) across subcategories.} Box plots display raw scores of each model.}
        \label{figure:subcategory_difficulty}
    \end{minipage}
    \hfill % This adds a small space between the two figures
    % Right table
    \begin{minipage}[t]{0.45\linewidth}
        \centering
        \includegraphics[width=\linewidth]{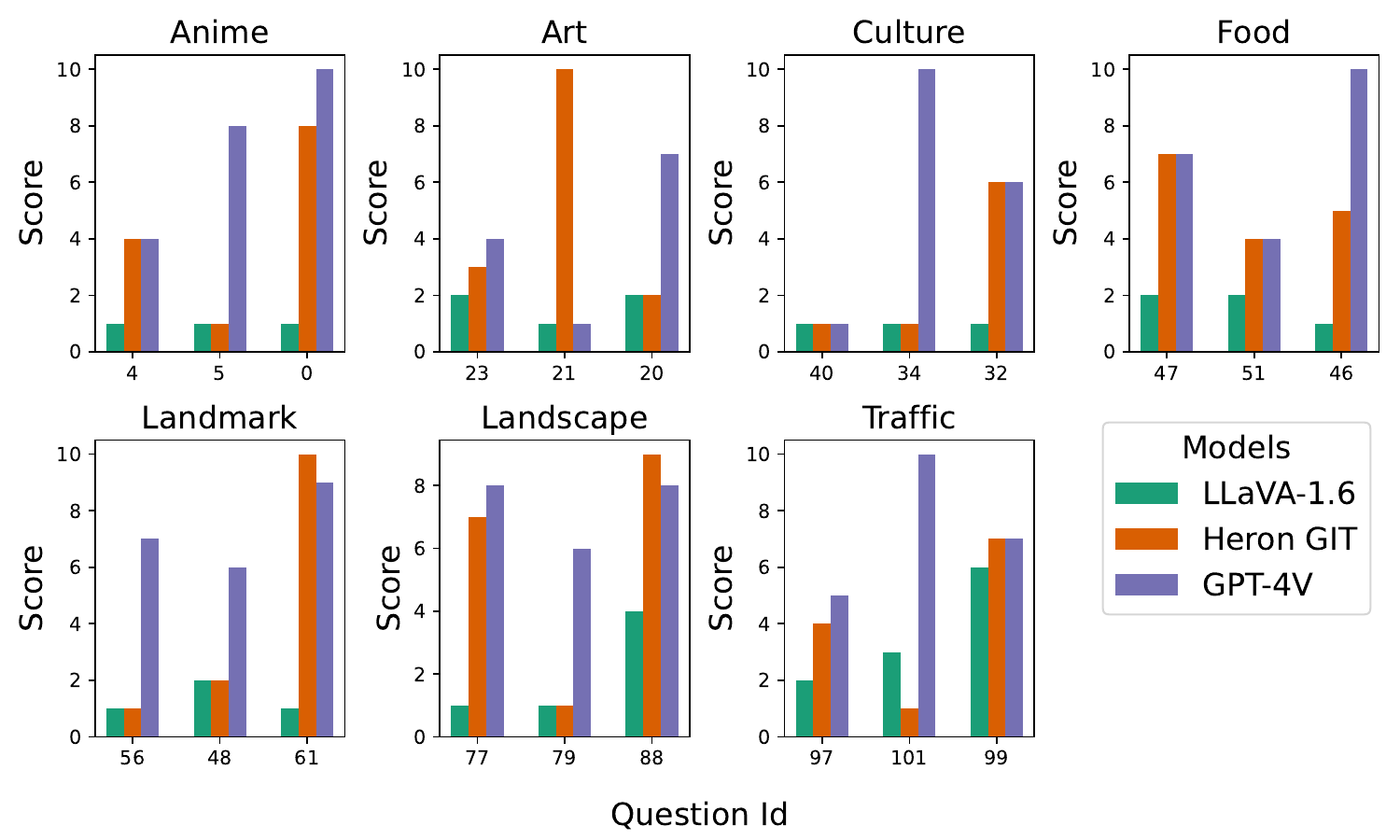} % Adjust the file name as needed
        \caption{\textbf{Comparison of scores for representative questions in each category.} Raw scores of GPT-4V, Heron GIT, and LLaVA-1.6 for three representative questions from each category are shown.}
        \label{figure:each_question_difficulty}
    \end{minipage}

\end{figure}

\subsection{Subcategories Analysis}

Figure~\ref{figure:subcategory_difficulty} shows the scores of GPT-4V (closed model), Heron GIT (Japanese VLM), and LLaVA-1.6 (English VLM) for each subcategory. Similar to the overall scores, GPT-4V exhibits high performance across all categories. Among the open models, each model has its strengths and weaknesses in different subcategories. Some subcategories, such as Traffic and Culture, have similar scores across the two models, while in others like Landmark, Food, Landscape, Art, and Anime, Heron GIT achieves higher scores.

Figure~\ref{figure:each_question_difficulty} presents raw scores of each model for three representative questions from each category. Examining the scores for individual questions reveals that each category contains questions of varying difficulty. Presence of questions with low scores, even for GPT-4V, in all categories suggests that there is room for evaluating models with even higher performance. While Heron GIT achieves high scores for some questions, GPT-4V consistently demonstrates high performance.

\subsection{Scoring Reproducibility}

\begin{figure}[H]
    % Left figure
    \begin{minipage}[t]{0.55\linewidth}
        \centering
        \includegraphics[width=\linewidth]{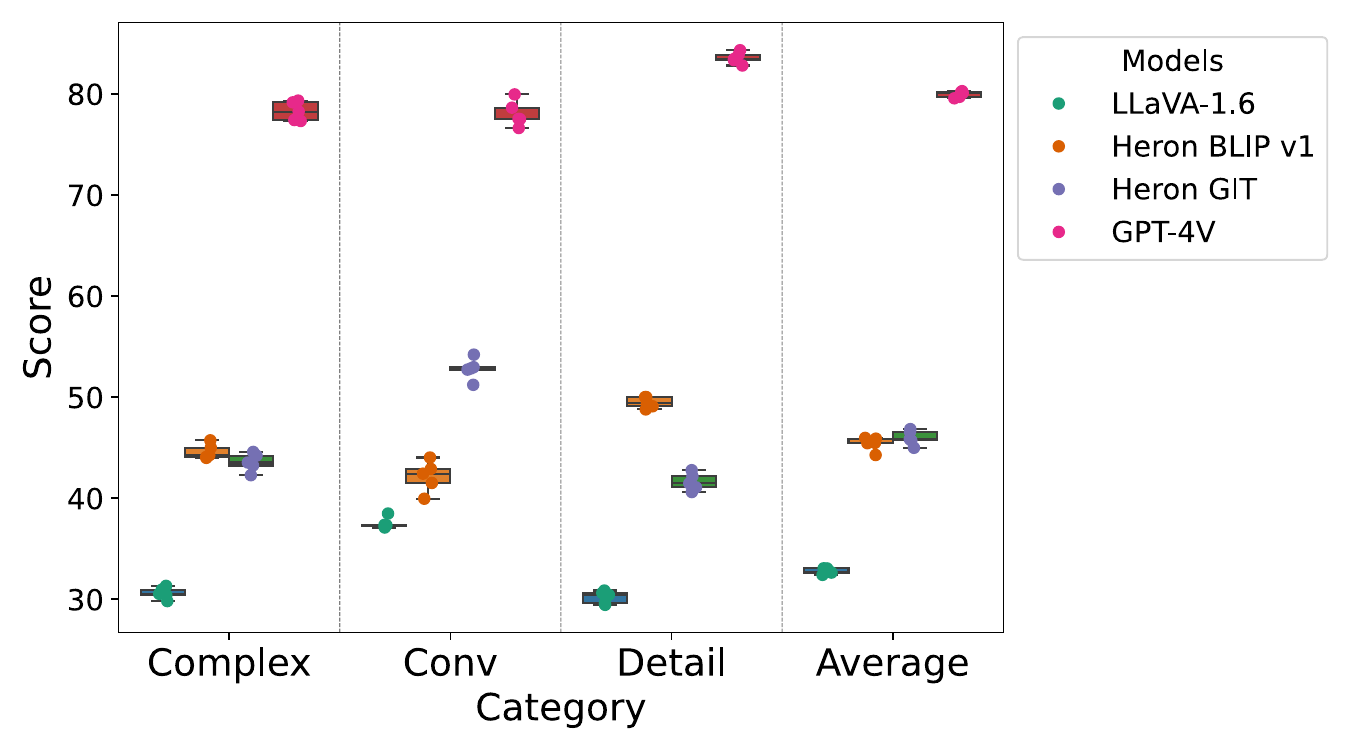} % Assuming the content of the table is an image
        \caption{\textbf{Variability in scores across five GPT-4 API calls for each model.} Bars represent average scores. Individual scores are also shown. Multiple evaluations can provide more accurate results when average scores between models are close.}
        \label{figure:fig_reproducibility}
    \end{minipage}
    \hfill % This adds a small space between the two figures
    % Right table
    \begin{minipage}[t]{0.43\linewidth}
        \centering
        \includegraphics[width=\linewidth]{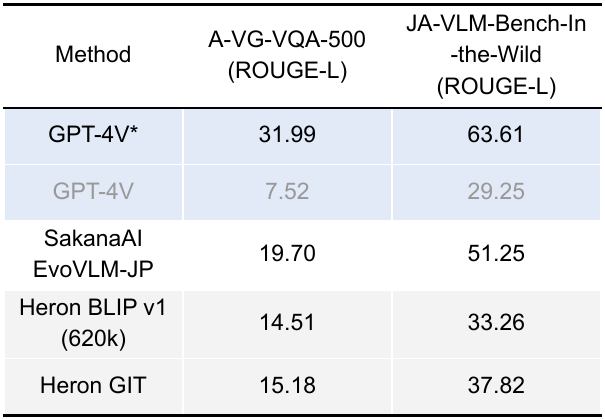} % Adjust the file name as needed
        \captionof{table}{\textbf{Comparison of the VLMs using JA-VG-VQA-500 and JA-VLM-Bench-In-th-Wild.} Asterisk (*) indicates that additional prompts were used to force the short answer.}
        \label{table:table_sakana_vqa}
    \end{minipage}

\end{figure}

API calls to GPT-4 do not yield deterministic responses, even when specifying configurations such as temperature and seed. In other words, when conducting evaluations using the GPT-4 API, complete reproducibility may not be achievable. Figure~\ref{figure:fig_reproducibility} shows the variability in scores when evaluating GPT-4 multiple times. We sent five requests to the GPT-4 API with temperature = 0 and seed = 0, and the results are presented. Looking at Complex, Conv, and Detail, we can see that although there is some variability, it falls within an acceptable range. Regarding the average scores, the variability within each model is not relatively significant. However, when the gap in average scores between models is around 1, as in the case of Heron BLIP v1 and Heron GIT, obtaining multiple evaluation results might provide more precise scores.

\subsection{Comparison with Existing Benchmarks}

JA-VG-VQA-500 and JA-VLM-Bench-In-the-Wild are available as evaluation metrics for Japanese VLMs~\cite{akiba2024evolutionary}. GPT-4V achieves high scores on these benchmarks, similar to its performance on our benchmark. However, the scores of the open models differ, with EvoVLM-JP scoring higher than Heron BLIP v1 and Heron GIT, suggesting that these benchmarks measure different aspects of model performance compared to ours. While our Heron-Bench uses GPT-4 for scoring, JA-VG-VQA-500 and JA-VLM-Bench-In-the-Wild employ ROUGE-L, indicating the different nature of scoring. We believe that our proposed Heron-Bench serves as a valuable new option for evaluating VLMs using images and questions that incorporate Japanese context.

% =============================================================
% Two tables side by side, Sakana Metrics and Sub categories
% =============================================================
% \begin{figure}[t!]

\subsection{Qualitative Evaluation}
\label{sec:qualitative}

We conducted a qualitative evaluation based on the Heron-Bench results. Tables~\ref{tab:qualitative_0}, \ref{tab:qualitative_1}, and \ref{tab:qualitative_2} show answer examples generated by Heron GIT, GPT-4V, and Claude 3 Opus for the Heron-Bench questions.

For the simple "Conversation" question shown in Table~\ref{tab:qualitative_0}, Heron GIT demonstrated that its answering capabilities are comparable to those of GPT-4V and Claude 3 Opus. However, it can be observed that Claude 3 Opus also provided unnecessary information in its answer. This suggests that further improvements are needed to provide the necessary and sufficient answers.

Table~\ref{tab:qualitative_1} presents a "Detail" question asking for an explanation of an image depicting a sumo's ring-entering ceremony called yokozuna's dohyo-iri (横綱土俵入り) and the corresponding answers. Despite the image showing three sumo wrestlers, all of the answers stated that "two sumo wrestlers are competing." This result suggests that the models' answers are influenced by the common knowledge that sumo matches typically involve two wrestlers. It implies that further improvements are necessary for the models to accurately interpret and convey the specific information captured in the image.

In the answers to a "Complex" question in Table~\ref{tab:qualitative_2}, all of the models succeeded in making the decision to stop. However, only GPT-4V correctly understood both the traffic light and the instructions given by the traffic guides. Consequently, GPT-4V provided a highly accurate response to the question by thoroughly understanding the context and incorporating the essential visual information.

\section{Conclusion}

In this work, we presented the Japanese Heron-Bench, a novel benchmark for evaluating the Japanese language capabilities of Vision Language Models. By constructing a diverse set of image-question-answer pairs relevant to the Japanese context, our benchmark enables a more comprehensive and culturally aware evaluation of VLMs in their Japanese language abilities. We also introduced a baseline Japanese VLM, Heron GIT, which employs the visual instruction tuning technique and achieves competitive performance compared to existing Japanese VLMs. We hope that the release of the Japanese Heron-Bench and our baseline model will encourage further developments in this area, leading to more capable and culturally aware Japanese VLMs.

\section{Limitations}

Although we have proposed the Japanese Heron-Bench as a dataset that encompasses Japanese cultural context, this evaluation method still has some limitations. Since the scores provided by GPT-4 are dependent on its own model performance, it may generate inaccurate responses to questions that exceed its knowledge. Particularly, as GPT-4's Japanese language performance is inferior compared to English, the scoring may be less precise when evaluating Japanese sentences compared to English ones. Furthermore, this method does not address the evaluation of safety aspects, and there is a possibility of generating misinformation, bias, hatefulness, or toxic content. To properly evaluate Japanese VLMs, further exploration of evaluation metrics is necessary.

\section*{Acknowledgments}

Part of computational resource of AI Bridging Cloud Infrastructure (ABCI) was awarded by “ABCI Grand Challenge” Program, National Institute of Advanced Industrial Science and Technology (AIST).

\bigskip
\bigskip
\bigskip
\bigskip
\bigskip
\bigskip
\bigskip
\bigskip
\bigskip
\bigskip
\bigskip
\bigskip
\bigskip
% \bigskip
% \bigskip
% \bigskip

%画像1. GPT-4V, Claudeは256の出力に修正済み
% \begin{table}[h]
\begin{table}[h]
    \centering
    {\large \textbf{Heron-Bench (Conversation): Anime}}
    \begin{tabular}{@{}p{6cm}p{10cm}@{}}
    \midrule
    \multicolumn{2}{c}{\includegraphics[width=8cm]{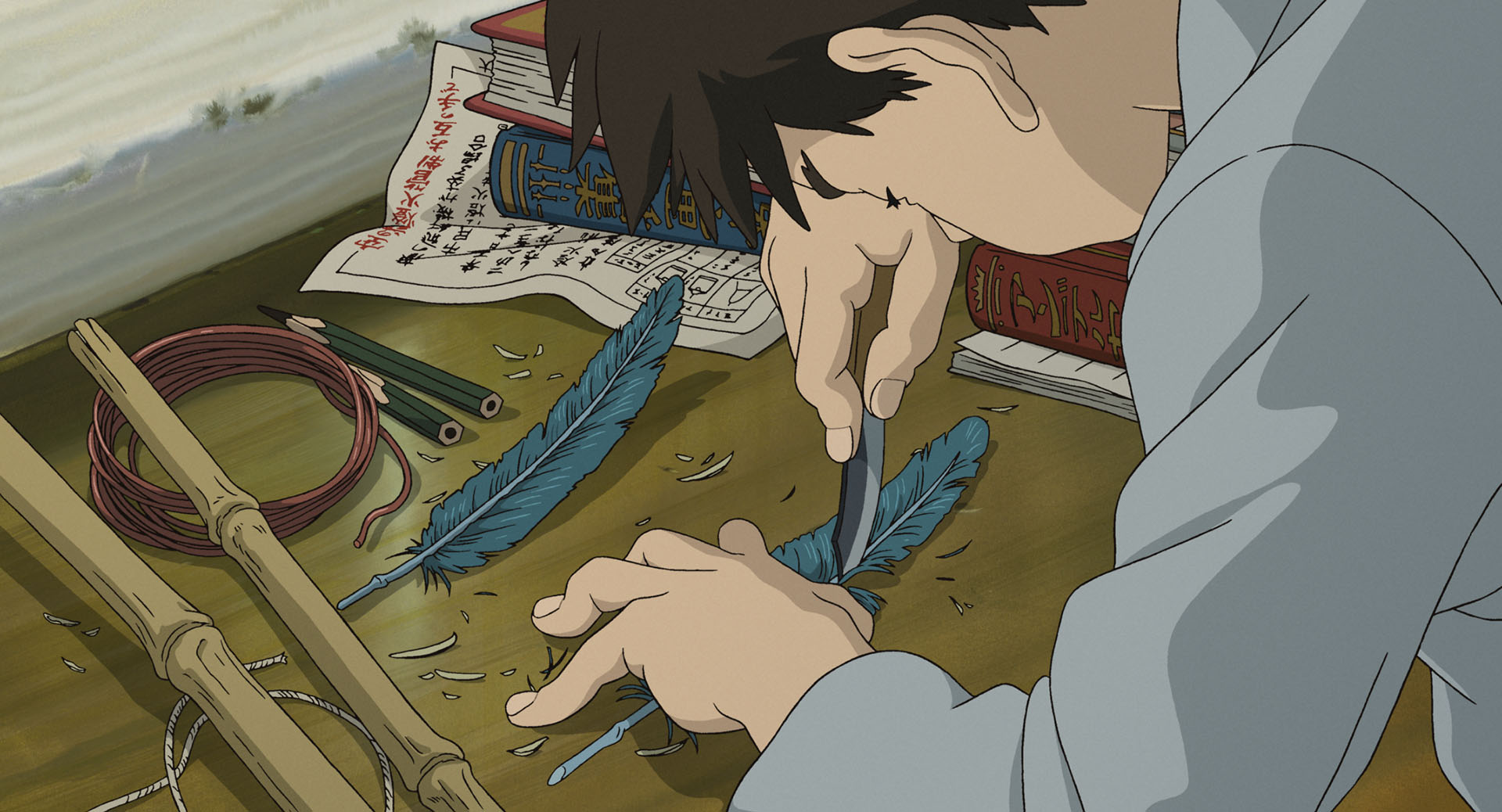}} \\
    % \midrule
    \textbf{User} & 鳥の羽の色は何色でしょうか？
    
    (What color are the bird's feathers?)\\
    \midrule
    \textbf{Heron GIT} & この鳥の羽の色は\underline{青}です。
    
    (The feather color of this bird is \underline{blue}.)\\ \\
    \textbf{GPT-4V} & この画像に描かれている鳥の羽の色は\underline{青色}です。

    (The color of the bird's feathers depicted in this image are \underline{blue}.) \\ \\
    \textbf{Claude 3 Opus} & 鳥の羽の色は\underline{青色}です。画像では、青い羽根ペンを持っている人物の手が描かれています。

    (The bird's feathers are \underline{blue}. In the image, a person's hand holding a blue feather pen is depicted.)\\
    \bottomrule
    \end{tabular}
    \caption{Qualitative evaluation for "Conversation" of VLMs.}
    \label{tab:qualitative_0}
\end{table}

\newpage

%画像2. GPT-4V, Claudeは256の出力に修正済み
% \begin{table}[h]
\begin{table}[H]
    \centering
    {\large \textbf{Heron-Bench (Detail): Culture}}
    \begin{tabular}{@{}p{6cm}p{10cm}@{}}
    \toprule
    \multicolumn{2}{c}{\includegraphics[width=6cm]{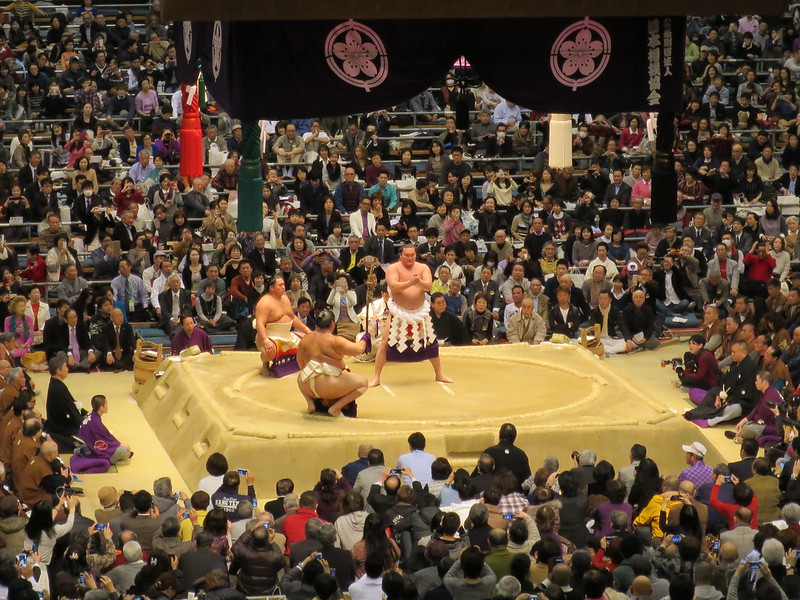}} \\
    \textbf{User} & 力士たちは何を行っているか説明してください。
    
    (Please describe what the sumo wrestlers are doing.)\\
    \midrule
    \textbf{Heron GIT} & 画像では、\underline{2人の力士のパフォーマンス}が行われており、1人はリングの上で相撲の技を披露し、もう1人の力士はリングの近くで太鼓を叩いています。このイベントは、相撲と太鼓のパフォーマンスを組み合わせたもので、観客を楽しませ、魅了するエンターテイメントとして機能します。
    
    (In the image, a \underline{performance by two sumo wrestlers} is taking place. One wrestler is demonstrating sumo techniques on the ring, while the other is playing a drum near the ring. This event combines sumo and drumming performances, functioning as entertainment to amuse and captivate the audience.)\\ \\
    \textbf{GPT-4V} & この写真は、日本の伝統的なスポーツである相撲の試合を捉えたものです。力士たちは土俵の上で対戦しており、周囲には観客が詰めかけています。写真の中央にいる\underline{二人の力士}は、試合の一環として、おそらく立ち合い直前の儀式を行っているところです。力士たちは特徴的なまわしを身につけており、試合に集中している様子が伺えます。土俵の周りには審判や他の関係者が見守っています。

    (This photo captures a match of sumo, a traditional Japanese sport. The wrestlers are competing on the dohyo (sumo ring), surrounded by a packed audience. \underline{The two wrestlers} in the center of the photo are likely performing a ritual just before the face-off, as part of the match. The wrestlers are wearing the distinctive mawashi (sumo belt) and appear to be focused on the match. Around the dohyo, referees and other officials are watching over the proceedings.)\\ \\
    \textbf{Claude 3 Opus} & 画像は相撲の取組の様子を捉えたものです。土俵の上で\underline{2人の力士が対峙}し、四股を踏んでいる最中のようです。土俵の周りには大勢の観客が見守っており、伝統的な相撲の興行の雰囲気が伝わってきます。力士たちは精一杯の力を出し合い、勝負の行方に観客も注目しているのが分かります。日本の国技である相撲の迫力ある一場面が写し出されています。

    (The image captures a sumo wrestling match. On the dohyo (sumo ring), \underline{two wrestlers are facing each other}, appearing to be in the midst of performing the shiko (leg-stomping ritual). A large audience surrounds the ring, watching intently, conveying the atmosphere of a traditional sumo event. The wrestlers are exerting their full strength against each other, with the audience keenly focused on the outcome of the match. A powerful moment of sumo, Japan's national sport, is depicted.)\\
    \bottomrule
    \end{tabular}
    \caption{Qualitative evaluation for "Detail" of VLMs.}
    \label{tab:qualitative_1}
\end{table}

%画像3. GPT-4V, Claudeは256の出力に修正済み
% \begin{table}[h]
\begin{table}[H]
    \centering
    {\large \textbf{Heron-Bench (Complex): Traffic}}
    \begin{tabular}{@{}p{6cm}p{10cm}@{}}
    \midrule
    \multicolumn{2}{c}{\includegraphics[width=8cm]{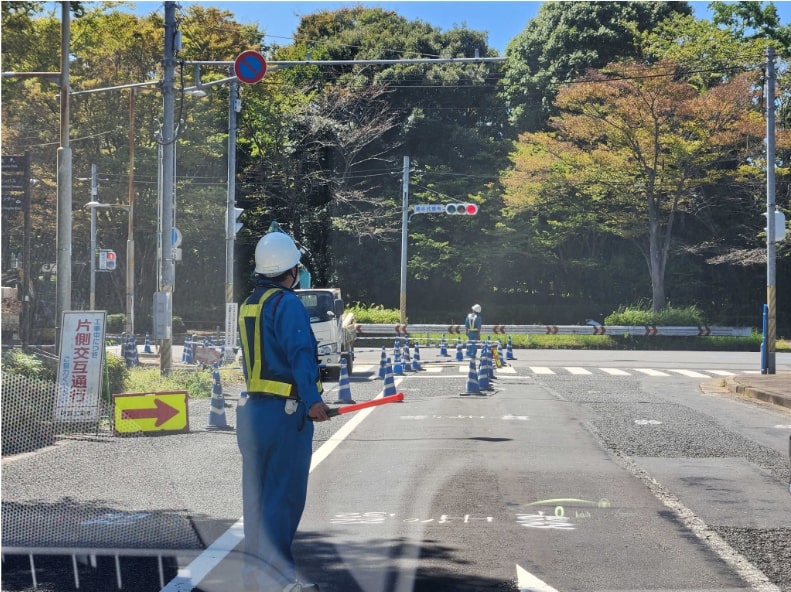}} \\ \\
    \textbf{User} & あなたは車を運転しており、交差点で右折しようと思っています。このまま交差点に向かってよいでしょうか？
    
    (You are driving a car and thinking about making a right turn at the intersection. Is it okay to continue towards the intersection?)\\
    \midrule
    \textbf{Heron GIT} & いいえ、あなたは右に曲がる前に、\underline{交通信号が変わるのを} \underline{待ってください。} 
    
    (No, please wait for the traffic light to change before you turn right.)\\ \\
    \textbf{GPT-4V} & 交差点に向かう前に停止する必要があります。画像には赤信号が表示されており、交通整理のために立っている作業員もいます。作業員は赤い旗を持っていて、交通を制御している様子が見受けられます。安全に右折するためには、\underline{信号と作業員の指示に従う}必要があります。信号が緑に変わり、作業員が進行を許可したときにのみ、交差点に進入して右折することができます。

    (You need to stop before approaching the intersection. The image shows a red light, and there is a worker standing for traffic control. The worker is holding a red flag, indicating that they are controlling traffic. To safely make a right turn, it is necessary to \underline{follow the signals and the instructions of the worker}. You can only enter the intersection and turn right when the signal turns green and the worker permits you to proceed.)\\ \\
    \textbf{Claude 3 Opus} & いいえ、このまま交差点に向かってはいけません。\underline{信号が赤}なので、一時停止して青信号になるまで待つ必要があります。交通ルールを守り、安全運転を心がけましょう。 

    (No, you must not proceed towards the intersection as it is. Since \underline{the light is red}, you need to come to a temporary stop and wait until it turns green. Let's adhere to traffic rules and focus on driving safely.)\\ \\
    \bottomrule
    \end{tabular}
    \caption{Qualitative evaluation for "Complex" of VLMs.}
    \label{tab:qualitative_2}
\end{table}

\newpage

%Bibliography
\bibliographystyle{unsrt}  
\bibliography{references}  

\newpage

\appendix

\section{Appendix: Prompt Design}
\label{appendix:a}

\begin{table}[H]
  \centering
  \begin{tabular}{@{}c@{}}
    \includegraphics[width=16cm]{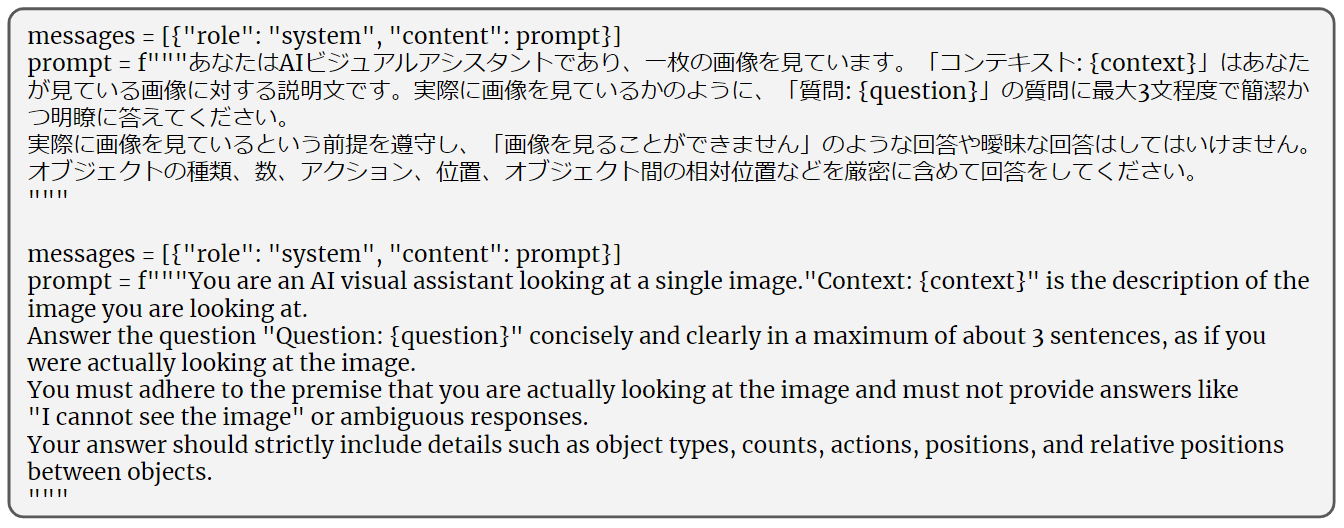} \\
  \end{tabular}
%   \caption{We illustrate the prompt used to generate responses from GPT-4. GPT-4 takes the context and question as input, and generates a response following the prompt.}
\end{table}

We illustrate the prompt used to generate answers from GPT-4. GPT-4 takes the context and question as input, and generates a response following the prompt. These answers are then used in GPT-4 scoring with VLM's answers.

\end{document}